\documentclass[10pt,journal]{IEEEtranTCOM}
\usepackage{times,amsmath,amssymb,graphicx,epsfig,color,colortbl}
\usepackage{tikz}
\usepackage{caption}
\usepackage{soul}
\usepackage[normalem]{ulem}
\usepackage{xcolor}
\usepackage{algorithm}
\usepackage[noend]{algpseudocode}

\makeatletter
\def\BState{\State\hskip-\ALG@thistlm}
\makeatother

\def\dist{\text{dist}}

\def\and{\text{and}}

\def\mb#1{\mathbf{#1}}

\def\nn{\nonumber}
\def\beq{\begin{equation}}
\def\eeq{\end{equation}}
\def\beqa{\begin{eqnarray}}
\def\eeqa{\end{eqnarray}}

\def\nn{\nonumber}

\def\beq{\begin{equation}}
\def\eeq{\end{equation}}
\def\beqa{\begin{eqnarray}}
\def\eeqa{\end{eqnarray}}

\def\bmtx{\begin{bmatrix}}
\def\emtx{\end{bmatrix}}

\begin{document}

\title{\bf Unbiased Sentence Encoder For Large-Scale Multi-lingual Search Engines}
\author{Mahdi Hajiaghayi, Mark Bolin and Monir Hajiaghayi $^{*}$\thanks{$^{*}$M
(c)2021 Microsoft Corporation. All rights reserved.  This document is provided "as-is." Information and views expressed in this document, including URL and other Internet Web site references (if any), may change without notice. You bear the risk of using it. This document does not provide you with any legal rights to any intellectual property in any Microsoft product.}\\
\IEEEauthorblockA{(c)Microsoft Corporation, Bellevue, WA 
\\E-mail: \{mahajiag, markbo, monirha\}@microsoft.com}}
\maketitle

\thispagestyle{empty}
\pagestyle{empty}

\begin{abstract}
In this paper, we present a multi-lingual sentence encoder that can be used in search engines as a query and document encoder. This embedding enables a semantic similarity score between queries and documents that can be an important feature in document ranking and relevancy. To train such a customized sentence encoder, it is beneficial to leverage users search data in the form of query-document clicked pairs however, we must avoid relying too much on search click data as it is biased and does not cover many unseen cases. The search data is heavily skewed towards short queries and for long queries is small and often noisy. The goal is to design a universal multi-lingual encoder that works for all cases and covers both short and long queries. We select a number of public NLI datasets in different languages and translation data and together with user search data we train a language model using a multi-task approach. A challenge is that these datasets are not homogeneous in terms of content, size and the balance ratio. While the public NLI datasets are usually two-sentence based with the same portion of positive and negative pairs, the user search data can contain multi-sentence documents and only positive pairs. We show how multi-task training enables us to leverage all these datasets and exploit knowledge sharing across these tasks.  
\end{abstract}

\noindent\textit{\textbf{Keywords \textemdash}} Multi-lingual Language Model, Universal Encoder, Deep Semantic Similarity, Multi-tasking.

\section{Introduction}

Sentence encoders have broad application in natural language processing (NLP) \cite{hu2020xtreme}-\nocite{conneau2017supervised}\cite{kiros2015skip} and information retrieval tasks \cite{mitra2017neural}. We focus on the ranking problem in large-scale search engines where for a given query, a retrieval model must return the most relevant documents from a large set of document candidates. Sentence encoding in such systems may be used to embed the query and documents independently into a common $N$-dimensional space such that the relevance score is found by the dot product of their representations. A retrieval system can precompute the embeddings of the document candidates and create an index map where, for an input query, the best matches can be obtained in sub-linear time (i.e., $O(Log(M))$ complexity where $M$ is the number of document candidates) \cite{ahmad2019reqa}. 

A popular approach to develop custom sentence encoders is to leverage a pre-trained language model and fine-tune it with user search data. This approach became popular after the advent of BERT \cite{devlin2018bert} in 2018 followed by other BERT-like models such as  XLNet \cite{yang2019xlnet}  RoBERTa \cite{liu2019roberta}, ALBERT \cite{lan2019albert} and ELECTRA \cite{clark2020electra}. The universal and multi-lingual versions of these works are available supporting more than 100 languages. These remarkable models are transformer-based and trained on massive text data using unsupervised objectives. 

The problem with fine-tuning using only user search data is that search data is often heavily biased.  The known biases include historical bias, position  bias and selection biases \cite{agarwal2019estimating}. For example, some search engines tend to give higher rank to the documents with high BM25 score.  This favors documents with more exact matching terms, and places them on the top of the rankings.  As a result, a semantically related document is less likely to appear in the search results and ultimately the derived training set. This in turn can bias the resulting language model to favor exact matches and give low similarity scores for semantic matches. This limits the model's ability to correctly generalize to future unseen cases. Similar unwanted behavior can result from a variety of other biases.

On the other hand, a general sentence encoder such as universal sentence encoders (USE) \cite{cer2018universal},\cite{yinfei2018learning} or SentenceBERT  \cite{reimers-2019-sentence-bert} or earlier works including InferSet \cite{conneau2017supervised} or Skip-Thought \cite{kiros2015skip} do not utilize user search data at all.  This limits their suitability for the search task. 

To address both problems mentioned above, we propose to use a multi-lingual multi-task approach that uses user search data together with natural language inference (NLI) datasets and translation data. Specifically, we use  SNLI \cite{snli:emnlp2015}, QNLI \cite{wang2018glue}, MNLI \cite{MNLI} , XNLI \cite{conneau2018xnli} and WikiMatrix \cite{wikimatrix}. As each of these datasets has its own characteristics such as size,  positive/negative ratio and content, the central question is how to combine these heterogeneous datasets. For example, NLI datasets are often balanced comprising the same ratio of positive and negative sentence pairs while user search data are only positive records containing queries consisting of a few words and matching documents with multiple sentences.

We define an appropriate loss function for each dataset that allows us to combine them in a multi-task setting. In particular, we consider cross entropy loss for the balanced datasets based on cosine similarity and a special form of the triplet loss for the user search data.   
 
Perhaps the closest architecture and work to our proposed model is SentenceBERT \cite{reimers-2019-sentence-bert} that uses a BERT model that is fine-tuned on NLI datasets. Since they use a single-task training approach, they provide multiple models for various scenarios such as semantic textual similarity or information retrieval. In our work, we provide a combined model that is able to outperform SentenceBERT in either of these scenarios, as will be shown in the experimental results. 

We will further discuss how each task must be scheduled and consumed for optimal performance. We also consider two approaches for representing a document with multiple text entities; attention-based and appending-based approaches. We will describe practical cases where the attention-based approach outperforms the appending-based in terms of better representing the document and achieving higher matching performance. Furthermore, as the majority of the pre-trained models are too big to be used in a practical search engine, we will review the various ways of knowledge distillation \cite{sanh2019distilbert} \cite{jiao2019tinybert} to reduce the complexity of these models. We will show that knowledge distillation while greatly reducing the model complexity, does not necessarily compromise the performance. In the experimental section, we provide comprehensive results and comparisons that show the benefits of our proposed model.
 
The rest of the paper is organized as follows. We first provide a brief description of sentence encoder usage in a Ranker in Section~\ref{Ranker}. In Section~\ref{mutli-tasking},  we describe our multi-task setup and various public datasets that together with user search data are used for training a multi-lingual sentence encoder. Furthermore, we formulate the training problem as an optimization problem in Section~\ref{loss_function} and discuss our choices for each task's loss function. The problem of scheduling multiple tasks is stated in Section~\ref{task_scheduling} and the proportional scheduling is introduced. We also consider two approach for representing a document with multiple text entities that are presented in Section\ref{concat_attn}. It is followed by a discussion in Section~\ref{base_encoder} on appropriate choice for the backbone of our model and how we can optimize and distill the structure to meet our ranking system requirements. Experimental results are given in Section~\ref{experimental_work}, followed by conclusion in Section\ref{conclusion}.

\section{Sentence Encoder in Ranker} \label{Ranker}

\begin{figure}[h]
 \centering
 \includegraphics[scale=.4]{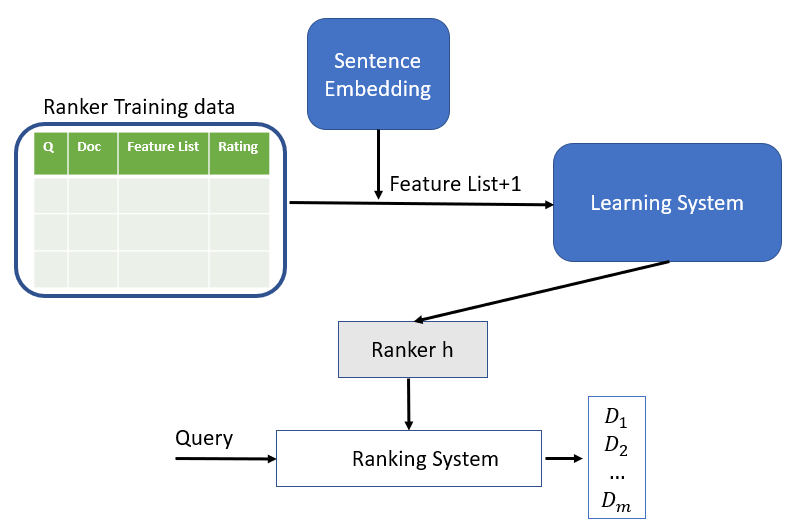}
 \caption{sentence encoder to produce a feature for stage $2$ ranker }
 \label{sentence_emb_ranker}
 \end{figure}

 A large-scale ranking or recommendation system often consists of multiple stages of ranking \cite{covington2016deep}. The first stage is often responsible for generating potential candidates from billions of documents while the other stages select the final candidates. A sentence encoder may be used to generate a semantic similarity score between a query and documents. This score can be used in all stages of ranking. The ranking mechanism in the first stage has to be simple and scalable to millions or billions of documents. The second stage often deals with a few thousands documents where a more advanced ranking algorithm such as 'learning to rank' \cite{liu2011learning} can be used. In the second stage where we get a few thousands candidates, a conventional ranker such as XGboost can be trained using the above relevancy score as a feature along with other query-document features. Figure \ref{sentence_emb_ranker} depicts how sentence encoder models can be applied to the query and documents columns on ranker training data and be appended to the data  as new additional features. Then the new modified data would be employed in a learn-to-rank mechanism in second stage of a ranker or recommendation system. 

\begin{figure}[h]
 \centering
 \includegraphics[scale=.35]{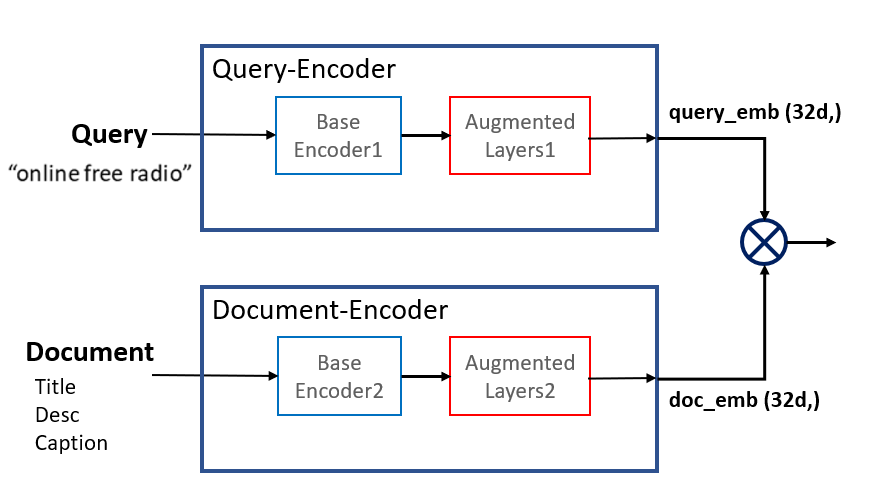}
 \caption{Dual Encoder model}
 \label{dual_encoder}
 \end{figure}

We consider a dual encoder framework that produces an embedding for query and documents as depicted in figure \ref{dual_encoder}. Each branch of this dual encoder has a base encoder which is often a BERT-like model followed by customized layers. This dual encoder takes a pair of query $q_i$ and documents $d_i$ and map them to $R^M$ space (M=32 or 100) $Em(q_i)$ and $Em(d_i)$ where a semantic similarity between them is obtained by dot product, i.e.  $Em(q_i)^T.Em(d_i)$. For now, we assume the same encoding function is applied to both query and documents but we will further relax this condition and argue that a practical system demands different encoder functions. 
The performance of a sentence encoder in rankers is evaluated through ranking metrics such as NDCG~\cite{yining2013theortical} or mean Average Precision~(mAP)~\cite{manning2009evaluation}. 

\section {Multi-Task approach: Combining  public datasets with user search data} \label{mutli-tasking}
The training data in a search engine is often collected from the user click and feedback data in the form of a positive pair $(q_i,d_i)$ where for a  given query $q_i$, document $d_i$ is reported to be related based on some custom metric. This metric can be based on the minimum number of clicks or click through rate, or watch time in the case of video documents.  For simplicity, we suppose we are only given the positive sets and the value of the metric such as actual click through rate (CTR) or number of clicks are not reported. 
\begin{table*}
\caption {Example of user search data}
\small
\begin{center}
\begin{tabular}{ | c || c | c | c || c | }
  \hline 
  & \multicolumn{3} {c|}{\textbf{Document}}
  & \bf{Language} 
   \\ \cline{2-4}
  \bf{Query} & \bf{Title} & \bf{Description} & \bf{Url} &  \\
  \hline \hline
  bumble bee sounds
  & The sound of Bumblebees
  & Nice sound effects in a bumblebee nest.
  & \shortstack{https://www.youtube.com/\\watch?v=XUQCO23C2eA}
  & En  \\   \hline
    happy wedding movie songs
  & The sound of Bumblebees
  & \shortstack{Play Free Music back to back only \\ on Eros Now - https://goo.gl/BEX4zD Listen \\ to all the Songs from Munna Michael  ...}    
  & \shortstack{https://www.youtube.com/\\watch?v=M4QepGzIkTs}
 & En \\  \hline

  chi quadrat test
  &  \shortstack{Chi2-Test:  Anpassungstest, \\Homogenitatstest und \\ Unabhangigkeitstest}

  & \shortstack{Der Chi2-Test funktioniert \\ auf jedem  Skalenniveau und \\ berücksichtigt immer ganze \\ Häufigkeitsverteilungen ...} 
  &  \shortstack{https://www.youtube.com/watch? ... \\  KurzesTutoriumStatistik}
 & De \\ \hline
       il etait une fois en amerique
  &  \shortstack{Il était une fois en Amérique \\ - Bande annonce vostFR \\- Sergio Leone}

  & \shortstack{Il était une fois en Amérique \\ -Bande annonce vostFR-... Film réalisé \\par Sergio Leone. Avec Robert De Niro ...}    
  & \shortstack{https://www.youtube.com/\\ watch?v=03N5qnZSWGE}
  & Fr \\ \hline

\end{tabular}
\label{data_snapshot_table}
\end{center}
\end{table*}

We also assume a multi-lingual retrieval system in which many languages are supported. As noted previously in many references, this data is heavily biased, and as a result, directly fine-tuning an embedding system on user search data does not necessarily add value to an existing ranker. Furthermore, if there was not an exploration phase in presenting the results, the performance of the sentence encoder and new ranker is bounded by the best performance of the previously shipped rankers. However, adding some public datasets to the training data used by our embedding system, allows some new and semantically related documents to surface on top results. 
To alleviate this issue, we propose to add some generic and unbiased datasets to our training data. In particular, we consider a set of natural language inference (NLI) datasets which involves inferring of whether two given sentences are entailed, neutral or contradicting each other.  We consider the following NLI datasets: 

\begin{enumerate}
\item SNLI (Stanford Natural Language Inference) \cite{snli:emnlp2015}: A collection of human-written English sentence pairs manually labeled for balanced classification with the labels entailment, contradiction, and neutral.
\item MNLI (Multi-Genre NLI): Similar to SNLI, but with a more diverse variety of text styles and topics, collected from transcribed speech, popular fiction, and government reports.
\item QNLI (Question NLI) \cite{wang2018glue}: Converted from SQuAD dataset to be a binary classification task over pairs of (question, sentence).
\item XNLI (Cross-Lingual NLI) \cite{conneau2018xnli} A multi-lingual extension of the SNLI/MultiNLI corpus in $15$ languages to evaluate such cross-lingual sentence encoder methods
\end{enumerate}
Inspired by \cite{yinfei2018learning}, we added the WikiMatrix \cite{wikimatrix} dataset and bridge tasks to align the embedding of different languages. WikiMatrix was extracted from wikipedia articles, and includes a complete set of parallel sentences across multiple languages. While the original dataset covers more than 93 languages and provides parallel corpora for each pair of these languages, we only used a subset of these massive datasets. We only considered 9 bi-lingual sentence pairs all aligned with English rendering around 5.7M pairs of sentences. Table~\ref{dataset_info_table} summarizes different datasets, along with  their language and size, used in our training data.

 This cross-lingual alignment task not only benefits the lower resource languages, but also serves to generate an embedding that is a suitable candidate for text to image matching, as vision has a universal embedding representation that is not language specific.
  
\begin{table*}
\caption {Multiple datasets used in training phase}
\small
\begin{center}
\begin{tabular}{ | c | c | c | }
  \hline
  \bf{DataSet} & \bf{Language} & \bf{Size} \\
  \hline \hline
  SNLI (Stanford NLI)
  & En
  & 360K
  \\   \hline
   MNLI(Multi-Genre NLI) 
  & En
  & 260K\\  \hline
       QNLI(Question Answer NLI)
  & En
  & 100K \\ \hline
   XNLI (multi-lingual)
  & 12 non-En languages inlcuding De-Fr-Es-Zh-Tr-Ar
  & 3.1M
  \\   \hline
   WikiMatrix
  & 
  En-Fr,En-Zh,En-Ja,En-It,De-En,En-Ru,En-Fa,En-Es,En-Ar
  & 5.7M\\  \hline
User Search Data 
  & 
  All Languages with higher portion for En, De, Fr, Es, It, Ja, Zh, Ko,Pr
  & 21M\\  \hline

\end{tabular}
\label{dataset_info_table}
\end{center}
\end{table*}

While the original forms of SNLI, MNLI and XNLI is 3-way with entailment, contradiction, and neutral, we ignore the contradiction portion to make them consistent with other datasets and user search data. Furthermore, this enables us to use a cosine similarity approach that has an efficient implementation for large-scaled datasets. 

\subsection{Loss Function} \label{loss_function}
The objective of our multi-task learning is to minimize the sum of the loss function for $T$ different tasks, through optimizing the network parameters. The objective function is therefore defined as 
\begin{equation}
\min_{\theta} \sum_{i=1}^T L_i(\theta,D_i), 
\end{equation} 
where $\theta$ is a set of the dual encoder parameters while $D_i$ and $L_i(.)$ denote the $i$-th task's dataset and loss function, respectively. The goal of multi-task learning is to find parameters $\theta$ and encoder function $f()$, which is embedded in the loss function, that achieve the best average performance across all the training tasks. Here we treat each sample pair as being equally important across all datasets. If size does not reflect the importance, the  loss of each dataset could be deferentially weighted.

For a balanced dataset with the same size of positive (entailment) and negative pairs (neutral), we use a cross entropy loss function \cite{kevin2012machine} for training. That is, given the ground-truth vector of outcomes $\mb{y} = [y_1, \cdots, y_N]$ and the vector of outcomes predicted by our model $\mb{\hat{y}} = [\hat{y}_1, \cdots, \hat{y}_N]$ for $N$ pairs of data, the cross-entropy loss function is calculated as 
\begin{align}
L_i(\theta, D = (\mb{q},\mb{d},\mb{y})) =\frac{1}{N} \sum_{j=1}^{N} y_j \log(\frac{1}{\hat{y}_j}) + (1-y_j) \log(\frac{1}{1-\hat{y}_j}).
\end{align}
where $\hat{y}_j = f(\theta, q_j)^T\dot f(\theta,d_j)$ in a dual encoder case. $f(\theta,x)$ represents the encoder function of value $x$. Here, we assume the outcome $\hat{y}_j$ lies in the $[0,1]$ interval which mandates either the embedding of both query and document are positive vectors or a projection of the cosine similarity onto the positive side, i.e.  $\hat{y}_j = max(0, f(\theta, q_i)^T\dot f(\theta,d_i))$.

For a dataset with only positive pairs we can generate negative pairs via random sampling. Specifically, corresponding to a positive pair of $(q_i,d_i)$, we randomly pick a document from the corpus to create a negative pair. Alternatively, we employ 'triplet loss' which was introduced to train an efficient face embedding for the task of face recognition \cite{schroff2015facenet}. Considering the usage of triplet loss function in our context, the idea is to minimize the distance between the embedding of query $q_j$ and document $d_j$, coming from a positive pair, while maximizing the distance between the embedding of query $q_j$ and negative documents $d_j^N$. The loss is therefore defined as 

\begin{align}
L_i(\theta, D = (\mb{q},\mb{d})) =  \frac{1}{N} \sum_{j=1}^{N} &\bigg [ \dist\left(f(\theta, q_j), f(\theta, d_j)\right) \nn \\  &- \dist\left(f(\theta, q_j) , f(\theta, d_j^N)\right)  + \alpha \bigg ]_+
\end{align}
 where $[x]_+  = \max(x, 0)$,$\alpha$ is a margin that is enforced between positive and negative pairs, and $\dist(x,y)$ can be any distance function such as cosine,  L1, or L2 distance. L1 distance, i.e. $\dist(x,y)=\|x-y\|$, works the best in our experiment setup. In practice, the negative document's embedding, $f(\theta, d_j^N)$ is computed in a mini-batch of stochastic gradient descent. Suppose we have a batch of $\mb B$ positive pairs of query and documents. There are different strategies to compute the negative document's embedding, out of which we have tried the following:
\begin{enumerate}
	\item batch-all triplet: The negative is mathematically calculated as 
	\beq
	f(\theta, d_j^N) = 1/|D| \sum_{d_k \in D d_k \neq d_j} f(\theta,d_k)  
	\eeq

where  
\begin{align}
D := \{d_k : &\dist\left(f(\theta, q_j), f(\theta, d_k)\right) \leq \nn \\
& \dist\left(f(\theta, q_j), f(\theta, d_j)\right) + \alpha \}
\end{align}
The intuition here is to discard all \it{easy} negative documents whose distance to query embedding (anchor) is greater than $\alpha$ plus the distance of the positive pair.  
	
	\item hard triplet: Rather than summing over all the negative samples, we can consider only the hard negative sample in a mini-batch of stochastic gradient descent. i.e. 
\beq
f(\theta, d_j^N) = \min_{d_k \in D, \: d_k \neq d_j} f(\theta,d_k)  
\eeq
	\item semi-hard triplet:  Selecting hard negative leads to bad convergence for some applications. Instead, 'semi-hard' negative chooses a hard negative that is not closer to the query embedding than the positive document embedding. i.e. 
\begin{align}
f(\theta, d_j^N) = & \min_{d_k \in D, \: d_k \neq d_j} f(\theta,d_k) \nn \\
& s.t.  \dist \left(f(\theta,q_j),f(\theta,d_j)\right) < \nn \\ 
& \dist \left(f(\theta,q_j),f(\theta,d_k)\right)
\end{align}
	\cite{schroff2015facenet} suggests semi-hard negative is shown to have better convergence performances in some cases. 
\end{enumerate}

We observed different performance of each of these strategies for different types of the base encoder that we use in our dual-encoder system. 

\subsection{Task Scheduling} \label{task_scheduling}
\begin{figure}[h]
 \centering
 \includegraphics[scale=.35]{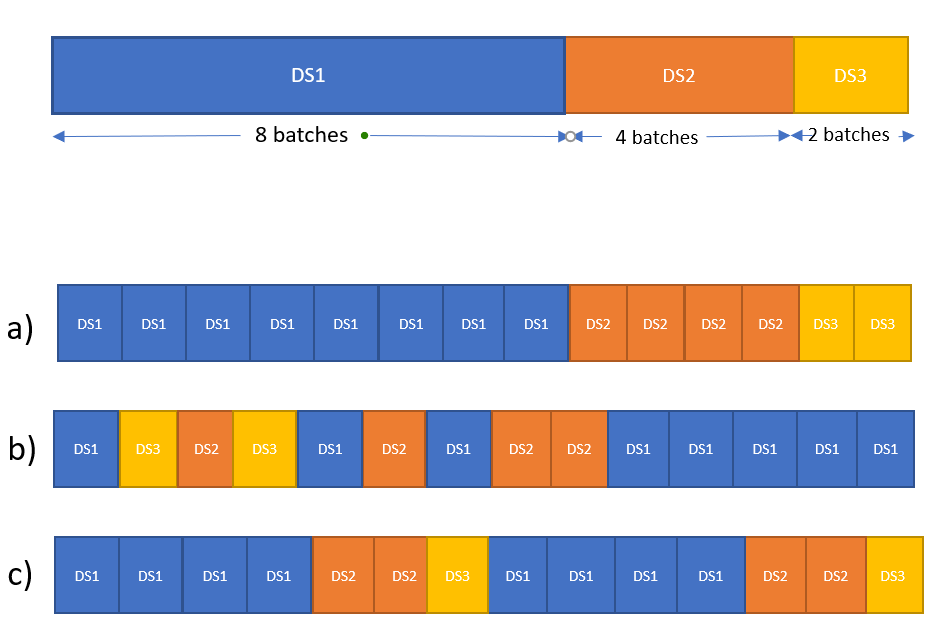}
 \caption{Illustration of different scheduling schemas for reading 3 datasets: a) Sequential read  b) Random read c) Proportional read}
 \label{proportoinal_read}
 \end{figure}

\begin{figure}[h]
 \centering
 \includegraphics[scale=.5]{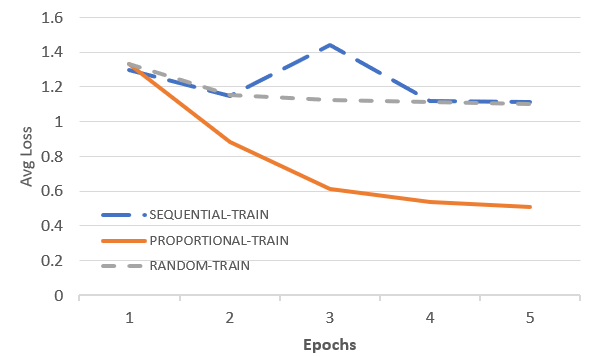}
 \caption{Performance comparison between various selection schemas}
 \label{graph_comp}
 \end{figure}

Multi-task learning
has emerged as a promising approach for sharing structure across multiple tasks to
enable more efficient learning \cite{yu2020gradient}. A task refers to a training dataset (NLI or other datasets) along with its loss function in our setup. A relevant question that may arise is that, in what order the datasets must be processed for training in order to achieve the highest performance with highest convergence speed?  A simple and easy-to-implement approach is to iterate through each dataset one by one and sequentially feed data to the model. This method that we call \textit{sequential read} is used in the work of Yang et al.~\cite{yinfei2018learning}. As each dataset may have different size, we found that the model trained with this approach may not fully utilize its capacity and may also converge very slowly. The other approach that we call \textit{random read}, would randomly pick a dataset and read a batch of data from the chosen dataset. We skip the dataset whose entire data has been processed in one epoch. In this approach, it is clear that the batches towards the end of each epoch comes from the larger datasets. 
The third and final approach is what we refer to as \textit{proportional read}. In this approach, we read batches from different datasets in a rate proportional to their sizes.  Fig \ref{proportoinal_read} depicts all these three approaches for a multi-task approach with three datasets with arbitrary sizes. A batch of data in this figure is represented by a block.
It is hard to predict which approach works the best in general, however, in our application we observed that the proportional read method significantly outperforms the other approaches. Fig. \ref{graph_comp} compares the average loss convergence speed for all these approaches over various iterations of training. Note that the loss function for all these three schemas is the same.  
\section{Document Embedding: Concatenation vs. Attention} \label{concat_attn}
A search query is typically a single sentence while a document may have multiple sources and entities of texts. For instance, a video document may contain title, description, closed caption and url. 
In order to provide an embedding for a document, we can concatenate all these text entities, creating a multi-sentence representation, and pass it to the encoder. This produces a single and fixed embedding for a document that can be used in a dot product operation to retrieve the most related documents for a given query in a large-scale system. 

\begin{figure}[h]
 \centering
 \includegraphics[scale=.7]{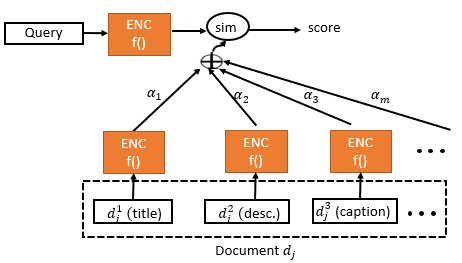}
 \caption{Attention approach to generate document embedding. Coefficients $\alpha_i$ is a function of text entity $d_j^i$ and given query $q$.}
 \label{fig_attention}
 \end{figure}

However, if we consider to use text embedding in the second or third stages of ranking where we have only a few thousands documents to rank, we can develop a better approach for utilizing all the text entities of a document. Specifically, one may consider \textit{attention} approach, which is very similar to the attention mechanism in  sequence and translation model~\cite{bahdanau2015neural}. However, the difference is that the attention in our case is in the sentence level and not word level. Here, the argument is that a document can have a dynamic embedding for different input queries. In other words, given a query, the attention mechanism decides to which part of the document's text (i.e., title, description, caption, etc) should attend more to generate a more meaningful document embedding. This procedure is depicted in Fig. \ref{fig_attention} where a document $d_j$ has $m$ separate text entities denoted as $d_j = (d^1_j,\cdots, d^m_j)$. The document embedding is obtained as weighted sum of the embeddings of each text entity, i.e, 

\begin{align}
f(\theta,d_j) = \sum_{k=1}^{m} \alpha_k f(\theta,d^k_j),
\end{align}
where the weights are output of a neural net with query and text entities as inputs followed by a softmax operation. Mathematically, $\alpha_k$ is yielded as 

\beq
\alpha_k = \frac{e^{a(q,d^k_j)}}{\sum_{l=1}^{m}  e^{a(q,d^l_j)}}
\eeq
and $a(x,y)$ can be a one-layer or multi-layers  neural net. 
 We conjecture the attention model is particularly useful when a text entity of the column is very noisy with rare useful information. For example, the URL of a video document does not often contains a meaningful information, unless some rare cases. 
\section{Base Encoder Selection for Query and Document Embedding} \label{base_encoder}
The backbone of our sentence encoder is a pre-trained language model. To be considered as a backbone, there are a few important characteristics that we are concerned from a language model. First, as our goal is to develop a multi-lingual sentence encoder, the base model must be multi-lingual and cover at least dozens most spoken languages. Secondly, we would like to cover the out of vocabulary (OOV) words.  This requirement is not met in a language model with word-level embedding. To address OOV cases, we need an embedding in a character level or sub-word level. The latter achieves a balance between the ﬂexibility of characters and efficiency of words. Sentence-piece \cite{kudo2018sentencepiece}  and Byte-Pair-Encoder (BEP) \cite{sennrich-etal-2016-neural} are two common approaches to generate such sub-word tokenizations. The last requirement is the size of the model and the inference time that should be strictly optimized.  In this study, we have considered multi-lingual base-BERT \cite{devlin2018bert},and Microsoft Turing model for natural language (NLR)~\cite{chi2020infoxlm}. To satisfy the third constraint, we prune and optimize the base models that we will discuss it in the next section.

\subsection{Encoder Optimization and Knowledge Distillation}
Due to the tight operation time constraint in a search engine, we must adopt a backbone architecture with slow inference time. Original BERT model presents some challenges in terms of having large size, slow inference time, and complex pre-training
process. Many attempts have been done to extract a simpler sub-architecture that maintains the same performance of its predecessor such as BORT model \cite{de2020optimal},  TinyBERT \cite{jiao2019tinybert}, which comes in a $4$ and a $6$-layers variant, and pruning transformer heads presented in work by Voita et al.~\cite{voita2019analyzing}. 
For simplicity, we construct our sentence encoder with a pre-trained BERT model with its first $6$-layers as a back-bone and further fine-tune it through multi-task approach discussed in previous section. 

We have so far assumed that the query encoder and document encoder are identical. However, They can be  separately trained with a specific setup based on their corresponding data, and then would be more  beneficial to a real-time retrieval system. This is because we have more flexibility with the document encoder as we can pre-calculate the embedding of our indexed documents, while it may not be the case for the live search queries. To mitigate this problem, we adopted the knowledge distillation technique in \cite{sanh2019distilbert} and \cite{jiao2019tinybert} to further reduce the complexity of the model and trained a 3-layers BERT model as a student model from a 6-layerss BERT model as a teacher. For training, we only used L2 loss between the output of the teacher and student encoders.

\section{Experiments} \label{experimental_work}
In this section, we evaluate the performance of our trained sentence encoders and compare them with some classical information retrieval methods and open source sentence encoders. 

\subsection{Training Setup}
We use the following settings for training. The maximum sentence length for document was set to $75$ tokens while that was set to $16$ tokens for the query part. After doing some experiments, we finalized the embedding dimensions to $M=32$. We also employ $6$-layers BERT for document embedding and $3$-layers BERT for query embedding. The $3$-layers BERT is called student model as it was trained from a 6-layerss teacher model. 
\subsection{Evaluation Datasets}
The main evaluation dataset in our case is a multi-lingual relevance dataset that is labeled by designated judges. The performance is reported as DCG score. This set has around $11,000$ distinct queries and more than 7M pair of query and documents with the given label. The difference between this set and user search data is that the effect of multiple bias terms is eliminated as the expert judges evaluate the relevancy of a document for a  given query. We further evaluate our model on public  XNLI and SNLI test sets, as one of our goal is to have a generalized sentence encoder that is not only bounded by the existing ranker's results but also performs rigorously on unseen data.  
  
\subsection{Experimental Results} \label{results}
We provide the results of running our sentence encoder against multiple data sets in table \ref{performance_results_table}. Specifically, we present the DCG scores at various depths of ranking for our multi-task multi-lingual $6$-layers BERT. To have a comprehensive comparison, we also evaluate the performance of some other models including the one without triplet loss, or single-task.

\definecolor{LightCyan}{rgb}{0.88,1,1}
\definecolor{blueColor}{rgb}{0.7,0.87,1}
\definecolor{greenColor}{rgb}{0.7,0.85,0.8}
\begin{table*}
\caption {Performance comparison of various sentence encoder algorithms.  Light cyan rows show the performance of optimal and simple models on our evaluation sets and are considered as our baselines to compare with. Blue lines present the results of off-the-shelf transformer models on our evaluation sets and the green lines indicate the performance of our proposed models.}
\label{table-results}
\begin{center}
\begin{tabular}{|c||c|c|c||c||c|} \hline
           & \multicolumn{3} {c|}{\textbf{Single-Feature Image DCG}} & \textbf{SNLI Val.} &\textbf{XNLI Val.}   \\ \cline{2-4}
\textbf{Models}  & @1  & @5 & @10 & \textbf{RoC} & \textbf{RoC}\\ \hline \hline
\rowcolor{LightCyan}
Random Algorithm          & 13.8  & 13.6  & 13.7   &  - & -   \\ \hline
\rowcolor{LightCyan}
Optimal          & 90.97  & 89.07  & 87.5    &  - & -  \\ \hline
\rowcolor{LightCyan}
Modified BM25         & 60.86    & 60.74  & 60.46  &  - & -     \\ \hline
\rowcolor{LightCyan}
Minimum Edit Distance \cite{levenshtein1966binary}        & 30.87 & 24.6  & 21.86   & - & -     \\ \hline
\rowcolor{blueColor}
USE 4.0 (512d)       & 56.4 & 56.0  & 55.6 & .71  & .57       \\ \hline
\rowcolor{blueColor}
USE 5.0 (512d)       & 59.15& 58.8  & 57.4 & .77  & .60       \\ \hline
\rowcolor{blueColor}
XLING-Large (512d)    & 60.05 & 60.38  & 60.42 & .78  & .68        \\ \hline
\rowcolor{blueColor}
SENTENCEBERT-Info. Retrieval (768d)         & 58.98  & 54.83   &  47.06    &  .72 & .57     \\ \hline
\rowcolor{blueColor}
SENTENCEBERT-Seman. Similarity (768d)    & 54.66    & 55.33  &  55.65   &  .74  & .70     \\ \hline
\rowcolor{blueColor}
BERT 12-layer (768d)     & 25.6   & 23.0  &  21.03    & .52  & .54     \\ \hline
\rowcolor{greenColor}
BERT 6-layers Multi-task (32d)     & \textbf{62.8}  &  62.5 &  62.3     & \textbf{.95}  & \textbf{.78}     \\ \hline
\rowcolor{greenColor}
BERT 6-layers Multi-task with Rand. Neg (32d)   & 60.21  & 60.0   & 59.5  & .92  & .76   \\ \hline
\rowcolor{greenColor}
BERT 6-layers Single-task (clicked-data) (32d)    & 62.79    & \textbf{62.69}  & \textbf{62.38}  & .73  & .62   \\ \hline
\rowcolor{greenColor}
BERT 3-layer Multi-task (student) (32d)    & 62.4  & 62.3  & 62.01  & .94 & .77        \\ \hline
\rowcolor{greenColor}
NLR 6-layers Multi-task (32d)       & 62.04 &61.9 & 61.7  & .94  & .77       \\ \hline
\end{tabular}
\label{performance_results_table}
\end{center}
\end{table*}

\subsection{Comparison with Classical Information Retrieval Methods}
We consider two popular ranking methods; 
First, BM25 method \cite{robertson2009probabilistic} that is shown to present a strong baseline. The original BM25 implementation uses word-level terms. We modified the implementation to the token-based terms in order to accommodate symbol-based languages. The tokenizer and dictionary were borrowed from the multi-lingual BERT. Second, Levenshtein distance or Minimum Edit Distance \cite{levenshtein1966binary} which measures how two given sentences are similar in terms of the minimum number of string operations (deletion, insertion and substitution) needed to change one sentence to another. 

For a given query and document, both of these methods with their conventional implementations generate a score that can be used to rank the documents associated with a query. Since the generated score is a relative score for each query, it can not be directly applied to determine whether two given sentences are related or not. While there is a special way to define an embedding vector based on BM25 method (CIDEr metric in \cite{vedantam2015cider}),  no results for these two methods are provided for the SNLI and XNLI validation set.  

On the multi-lingual image relevance ranking dataset, as can be seen in table~\ref{table-results}, our proposed model 'BERT 6-layerss multi-task' significantly outperforms the Levenshtein distance's method with large margin and  BM25 method with less margin. BM25 method relies on exact matching and for a ranking dataset where query and document share many common tokens, BM25 is a difficult baseline to beat. A similar observation was made in \cite{lee2019latent} for some  question answering (QA) datasets.

\subsection{Comparison with Universal Sentence Encoders}
Next, we compare our models with off-the-shelf, generic sentence encoders. Specifically, we evaluate the performance of two versions (V4 and V5) of Universal Sentence Encoder (USE)\footnote{These Models are available at https://tfhub.dev/} \cite{cer2018universal} and XLING model \cite{yinfei2018learning} which is a multi-lingual sentence encoder that supports $16$ languages with $3$ layer transformer architecture and sentence-piece tokenization. The output of all these models is a $512$ dimensional vector.  Among all these models, XLING achieves the highest performance both on the single-feature ranking and NLI validation datasets. However, our proposed model demonstrates much better performance on SNLI and XNLI datasets while improving the relevancy score by $2.5$ DCG point on average over the various ranking depths.  

SentenceBERT \cite{reimers-2019-sentence-bert} also offers trained models for different tasks. One task is textual semantic similarity where a multilingual model is trained on NLI tasks. They also have a custom model for  information retrieval and ranking task based on MSMARCO dataset \cite{MSMARCO} but it is limited to the English Language. We evaluate both of these models in Table~\ref{table-results}. Our model outperforms both of these models on NLI and ranking tasks by a significant margins.

We also examined the performance of pre-trained BERT model without fine-tuning. From the table~\ref{table-results} it is evident that pre-trained model performs very poorly and needs fine-tuning for achieving better performance.

\subsection{Effect of Multi-Tasking and Triplet Loss}
We first compare the results of two models, both based on $6$-layers BERT as the backbone; one trained through multi-tasking on the datasets described in Section~\ref{mutli-tasking} and one solely trained on user search data. These two models are specified by 'BERT 6-layers Multi-task (32d)' and 'BERT 6-layers Single-task (clicked-data) (32d)' in Table~\ref{table-results}, respectively. It is remarkable to observe that while multi-tasking leads to no regression on the relevance dataset, it results in a significant boost in performance on SNLI and XNLI validation sets. Hence, multi-tasking here assures us that can best cover the unseen cases of query-documents and more robust with respect to position and selection biases in the clicked data. 

Second, we trained a model where we apply random negative strategy instead of semi-hard or hard negatives in the triplet loss. We observe that this model regresses by more than $2.0$ DCG points (from $62.8$ to $60.21$ at depth $@1$) indicating triplet loss and ignoring easy negative documents is a better choice.  

\section{Conclusion} \label{conclusion}
This paper has presented a multi-lingual sentence encoder that can be used in large scale search engines.  This encoder can be used to independently encode queries and documents.  This enables an inexpensive calculation of query to document similarity scores that can be used at various stages of ranking.
Training a sentence encoder requires large scale domain specific training data, such as can be derived from search user query-document click pairs.  However, this data can be heavily biased.  We have shown how multi-task training exploiting the public domain NLI and WikiMatrix datasets in conjunction with user search data can be used to reduce this bias and produce an encoder that is more robust to unseen queries.

\bibliographystyle{IEEEtran}
\bibliography{IEEEabrv,multitask_lm}
\end{document}